\documentclass[letterpaper, 10 pt, conference]{ieeeconf}  % Comment this line out if you need a4paper

\usepackage{amsmath,amssymb,amsfonts}
\usepackage{algorithmic}
\usepackage{graphicx}
\usepackage{textcomp}
\usepackage{xcolor}

\usepackage{mathtools}
\usepackage[caption=false]{subfig}
\usepackage{booktabs}

\usepackage[hidelinks]{hyperref}
\usepackage{cite}
\usepackage[capitalize,nameinlink]{cleveref}
\crefname{figure}{Fig.}{Figs.}
\Crefname{figure}{Fig.}{Figs.}

\begin{document}
\title{Learning Transferable Friction Models and LuGre Identification via Physics-Informed Neural Networks}
\author{Asutay Ozmen, Jo{\~a}o P. Hespanha, Katie Byl}%
\maketitle
\begin{abstract}
% Accurately modeling friction in robotics remains a core challenge, as robotics simulators like MuJoCo and PyBullet use simplified friction models or heuristics to balance computational efficiency with accuracy, where these simplifications and approximations can lead to substantial differences between simulated and physical performance. In this paper, we present a physics-informed friction estimation framework that enables the integration of well-established friction models with learnable components—requiring only minimal, generic measurement data. Our approach enforces physical consistency yet retains the flexibility to adapt to real-world complexities. We demonstrate, on an underactuated and nonlinear system, that the learned friction models, trained solely on small and noisy datasets, accurately simulate dynamic friction properties and reduce the sim-to-real gap. Crucially, we show that our approach enables the learned models to be transferable to systems they are not trained on. This ability to generalize across multiple systems streamlines friction modeling for complex, underactuated tasks, offering a scalable and interpretable path toward bridging the sim-to-real gap in robotics and control.
Accurately modeling friction in robotics remains a core challenge, as robotics simulators like MuJoCo and PyBullet use simplified friction models or heuristics to balance computational efficiency with accuracy, where these simplifications and approximations can lead to substantial differences between simulated and physical performance. In this paper, we present a physics-informed friction estimation framework that enables the integration of well-established friction models with learnable components, requiring only minimal, generic measurement data. Our approach enforces physical consistency yet retains the flexibility to capture complex friction phenomena. We demonstrate, on an underactuated and nonlinear system, that the learned friction models, trained solely on small and noisy datasets, accurately reproduce dynamic friction properties with significantly higher fidelity than the simplified models commonly used in robotics simulators. Crucially, we show that our approach enables the learned models to be transferable to systems they are not trained on. This ability to generalize across multiple systems streamlines friction modeling for complex, underactuated tasks, offering a scalable and interpretable path toward improving friction model accuracy in robotics and control.
\end{abstract}

% \begin{IEEEkeywords}
% Friction modeling, physics-informed neural networks, underactuated systems, sim-to-real gap, robotics, transferability
% \end{IEEEkeywords}
\section{Introduction} 
Machine learning and deep learning have gained momentum in solving complex problems in computer vision, natural language processing, and generative modeling. Large, high-quality datasets are often available in these domains or can be synthesized to train data-hungry models. For instance, virtual 3D environments provide large synthetic data for tasks like obstacle detection, mapping, and navigation in robotics~\cite{airSim,iGibson}. These data-driven strategies work well in purely digital domains or when high-fidelity virtual environments are available.

However, in real-world \emph{physical} interactions involving \emph{contact} and \emph{friction}, high-quality data is scarce. Real-world experiments are expensive, time-consuming, and subject to noise or wear-and-tear constraints that make large-scale data collection impractical. This challenge is amplified when robotic systems must deal with unpredictable or varying contact conditions, such as friction changes or impacts during locomotion and manipulation. In particular, underactuated systems that rely on friction for locomotion face challenges in modeling continuous sliding contacts accurately~\cite{contactModelsinSim}. Many robotics simulators, such as MuJoCo or PyBullet, compensate by using simplified friction models or heuristics to balance computational efficiency with accuracy~\cite{le2024contact}. Since friction is highly dependent on local surface properties, velocity, and normal force, these simplifications can lead to substantial differences between simulated and physical performance. This mismatch, commonly referred to as the \emph{sim-to-real} gap, can be especially pronounced in applications where friction plays a key role in system stability and control. To mitigate the resulting inaccuracies, methods such as Model Predictive Control (MPC) incorporate high-frequency feedback by repeatedly solving Optimal Control problems at rates up to 1~kHz~\cite{Diehl2006, mpc2014, nonlinearMPC,mpc2022}.

\subsection{Related Work}
Physics-Informed Neural Networks (PINNs) have found applications across a wide range of fields, including fluid dynamics, plasma physics, quantum chemistry, and material science~\cite{flowEspresso,plasmaPhys,quantumChem,structural,4DflowMRI}. PINNs have shown particular promise in friction modeling, structural hysteresis prediction, and robotic joint dynamics~\cite{li2024, coble2024, sorrentino2024}. They offer significant advantages such as data efficiency, interpretability, and computational efficiency, reducing dependency on extensive datasets and discretizations~\cite{raissi2019, karniadakis2021, li2024}. However, training complexity, scalability, and generalization remain open challenges, motivating ongoing research on hybrid models and optimization strategies~\cite{coble2024, karniadakis2021, li2024}.
% Overall, PINNs represent a paradigm shift in scientific computing, bridging the gap between machine learning and traditional physics-based modeling. Their ability to integrate domain knowledge into data-driven frameworks has unlocked new possibilities for solving complex problems in science and engineering. Hence, we leverage PINNs in this paper to learn the frictional properties of an environment where a system is deployed.

A few researchers have proposed learning-based friction models to bridge the \emph{sim-to-real} gap. Sorrentino~\textit{et al.} incorporate friction data \emph{a priori} into the training process, using measured friction forces to guide training losses~\cite{sorrentino2024}. Others omit direct friction force measurements but rely on simplified linear friction terms~\cite{scholl2024,olejnik2023}, which may be too restrictive to capture complex frictional phenomena, particularly for surface friction characteristics such as stick-slip behavior under time-varying normal forces or in underactuated settings. Moreover, large-scale data collection for frictional interactions is far more demanding than collecting images or text, limiting how effectively purely data-driven methods can scale.

More broadly, hybrid approaches that combine physics-based models with data-driven components have gained traction across dynamical systems modeling~\cite{karniadakis2021}. These methods range from using neural networks to learn residual corrections on top of known dynamics to embedding physical structure directly into the learning architecture. Our work falls into the latter category: rather than learning a correction term or using physics only as a regularizer, we embed the LuGre friction model structure into the network's forward pass, enabling the framework to learn both the friction behavior and the underlying model parameters from state data alone.

Decades of friction modeling research have yielded detailed models and theoretical insights, most notably advanced friction representations such as the LuGre model~\cite{canudas1995new}, which captures dynamic friction behaviors like stick-slip motion and the Stribeck effect. However, a hybrid approach that integrates well-established friction models into a learning-based framework remains underexplored. Our hybrid approach enables the learned friction models to gain interpretability and physical consistency while still leveraging data to correct for model inaccuracies in the face of real-world complexities.
\subsection{Aim of the Paper}

We propose a physics-informed learning approach that learns a friction model with minimal reliance on extensive friction datasets and incorporates learnings from prior research on friction modeling. Our methods use only the system's states and the governing equations of motion (EoMs). Building on ideas from PINNs, we embed the friction model within a loss term based on equations of motion, ensuring that friction estimates remain consistent with the physics that governs the system. This approach is versatile since it can be implemented as a black-box neural network or as a parameter estimation model that explicitly identifies the parameters of a known friction formulation. Rather than starting from scratch, our hybrid formulation leverages decades of research in friction models and provides additional benefits such as interpretability and potential parameter/model reuse across different systems with similar surface contact properties.

Our framework differs from standard PINN formulations in several key aspects. We drop the data loss term entirely: the network is trained solely through the EoM residual, with no direct supervision on the output quantity. The network learns to predict quantities that are never directly measured in the training data, namely friction force in the black-box models and the LuGre internal bristle state $z$ in the parameter estimation models. The physics loss is therefore not a regularizer augmenting data-driven learning, but the exclusive training signal for inferring unobserved physical quantities from readily available state measurements. Furthermore, in our parameter estimation models, the LuGre friction equations are architecturally embedded in the network's forward pass: the LuGre parameters ($\sigma_{0,1,2}$, $\mu_{c,s}$, $v_s$) are learned as trainable variables, and the friction force is computed through the LuGre structure rather than approximated by a generic function. This structural integration is reinforced by a consistency loss $\mathcal{L}_{\dot{z}}$ that couples the network's output, obtained via automatic differentiation, to the LuGre state evolution equation, ensuring that the learned internal state and parameters remain mutually consistent.

In contrast to prior learning-based friction methods that either require direct friction force measurements during training~\cite{sorrentino2024} or rely on simplified linear friction terms~\cite{scholl2024, olejnik2023}, our approach requires neither, learning complex dynamic friction behavior from generic state data alone. To the best of our knowledge, this is one of the few frameworks providing PINN-based transferable learned friction models that focus on surface friction rather than joint or drive friction. We demonstrate that models trained on one system generalize to a different dynamical system sharing the same contact surface, and that the parameter estimation variant identifies LuGre parameters with accuracy comparable to established methods such as Nelder--Mead and genetic algorithms while being significantly faster. Our aim is to show that even with limited and noisy datasets, high-fidelity friction models can be efficiently learned, offering improved friction model accuracy over the simplified models commonly used in robotics simulators.

The rest of the paper is structured as follows. In \cref{sec:background}, we outline the preliminaries in the form of the LuGre friction model and PINNs that are necessary for our framework. \Cref{sec:methods} outlines our PINN-based framework. \Cref{sec:results} shows that our framework is suitable for learning friction models that can be used as in-simulation friction models and online friction estimators, as well as being transferable to different dynamical systems.
\section{Preliminaries}\label{sec:background}
\subsection{LuGre Friction Model}\label{sec:lugre}
Introduced by Canudas de Wit~\textit{et al.} (1995), the LuGre friction model significantly advanced friction modeling by addressing the shortcomings of classical static friction models like Coulomb and viscous friction, particularly at low velocities and during velocity reversals~\cite{canudas1995new}.

Traditional models often neglect dynamic behaviors such as hysteresis and the Stribeck effect, where friction force decreases after a certain velocity threshold~\cite{marques2021}. The LuGre model captures these effects by introducing an internal state representing microscopic bristle deflection at the frictional interface. This state evolves with relative velocity, allowing simulation of pre-sliding displacement and varying break-away forces~\cite{astrom2008revisiting}.
% \subsubsection*{Mathematical Formulation}
\begin{table}[tp]
\caption{Parameters of LuGre friction model}
\label{lugrepar}
\centering
\footnotesize
\begin{tabular}{@{}ll@{}}
\toprule
\textbf{Parameter} & \textbf{Description} \\ \midrule
$\boldsymbol{\sigma_0}$ & Bristle stiffness \\
$\boldsymbol{\sigma_1}$ & Bristle damping \\
$\boldsymbol{\sigma_2}$ & Viscous damping coefficient \\
$\boldsymbol{F_s},\boldsymbol{\mu_s}$ & Static friction force and coefficient \\
$\boldsymbol{F_c},\boldsymbol{\mu_c}$ & Coulomb friction force and coefficient \\
$\boldsymbol{v_s}$ & Stribeck velocity \\
$\boldsymbol{\alpha}$ & Transition shape factor \\ \bottomrule
\end{tabular}
\end{table}
The LuGre friction model is characterized by the internal state evolution \eqref{eqn:zdot} and the linear friction force \eqref{eqn:Ffriction} shown below:
\begin{align} 
    \dot{z} &= v - \frac{\sigma_0 |v|}{F_c + (F_s - F_c) e^{-(|v|/v_s)^\alpha}} z  \label{eqn:zdot}\\
    F_f &= \sigma_0 z + \sigma_1 \dot{z} + \sigma_2 v \label{eqn:Ffriction}
\end{align}
Here, \( z \) represents the internal state variable corresponding to the average deflection of microscopic bristles at the contact interface, and \( v \) denotes the relative velocity between the surfaces. The denominator of the function that is multiplied by $z$ in \eqref{eqn:zdot} encapsulates the velocity-dependent nonlinear characteristics of friction, notably the Stribeck effect, which describes the reduction in friction force with increasing velocity past a certain threshold. In this paper, we represent Coulomb and static friction forces $F_c$ and $F_s$ in~\eqref{eqn:zdot} as functions of their respective friction coefficients $\mu_c$, $\mu_s$ and the normal force $F_N$. $F_f$ in~\eqref{eqn:Ffriction} is the resulting friction force, where the LuGre friction parameters are defined in \cref{lugrepar}.
\begin{align}
    F_c &= \mu_c|F_N| \label{eqn:Fc} \\
    F_s &= \mu_s|F_N|  \label{eqn:Fs} 
\end{align}

These parameters are typically determined through experimental identification methods, which involve measuring frictional forces under controlled conditions to fit the model accurately to observed behavior~\cite{cyrusID}. By appropriately selecting and calibrating these parameters, the LuGre model can effectively simulate various frictional phenomena, including stick-slip motion, hysteresis, and pre-sliding displacement, making it a valuable tool in the analysis and control of mechanical systems subject to friction~\cite{astrom2008revisiting}.
\begin{figure}[tp]
    \centering
    \includegraphics[width=0.5\columnwidth]{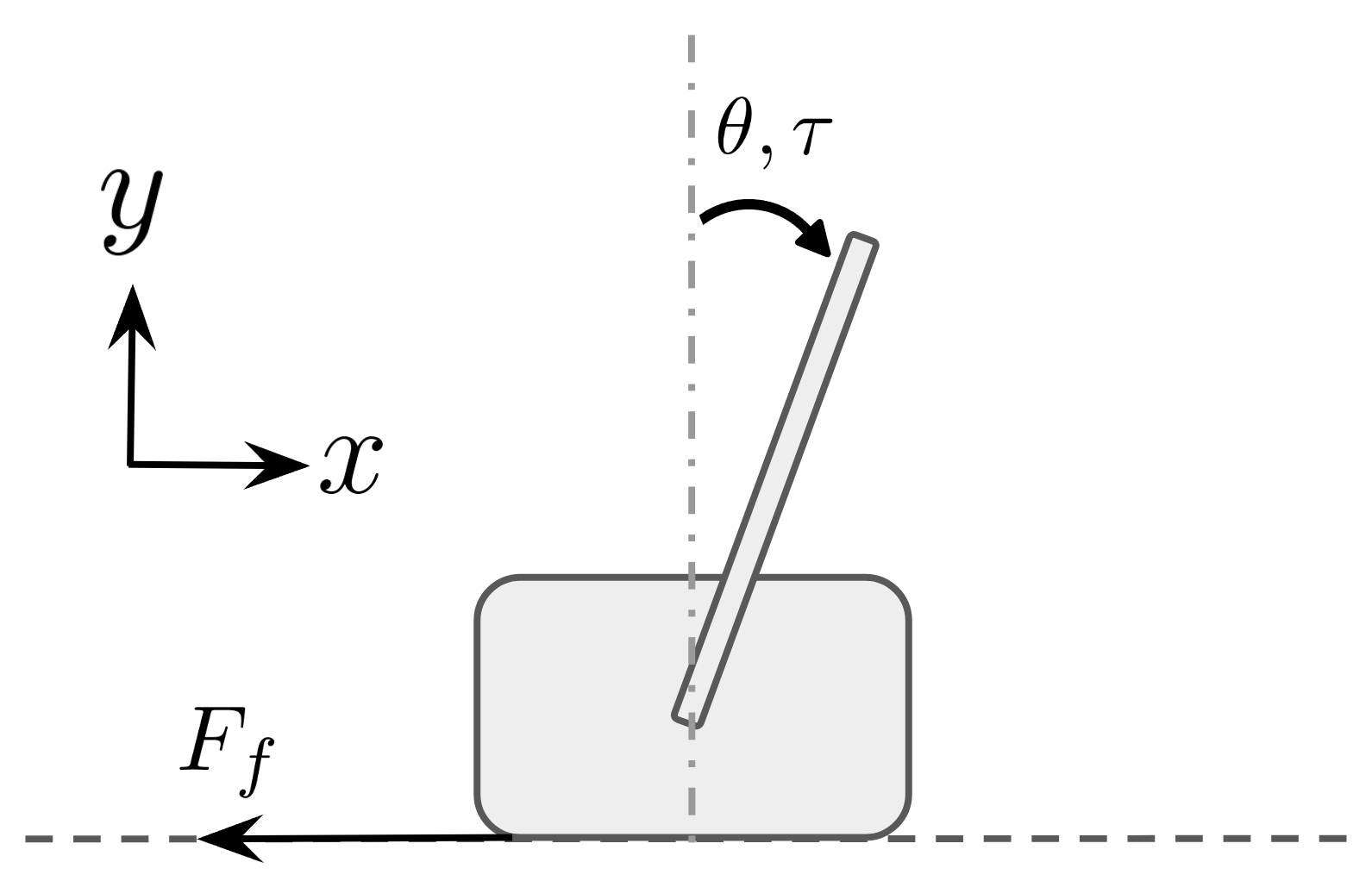}
    \caption{\footnotesize Pendulum-on-a-Box system}
    \label{fig:PoB}
\end{figure}
The LuGre friction model represents a foundational tool in the study and control of frictional systems, offering a dynamic framework that captures a wide range of frictional behaviors beyond the capabilities of traditional static models. Although some challenges, including drift and discrepancies in non-stationary regimes, continue to drive refinements and new friction models~\cite{rillLuGreNotLuGre2024}, for the purpose of this paper, we will be focusing on the base LuGre model defined by~\eqref{eqn:zdot} and~\eqref{eqn:Ffriction}.

\subsection{Physics-Informed Neural Networks (PINNs)} \label{sec:pinns}
Physics-Informed Neural Networks (PINNs) introduced by Raissi \textit{et al.} (2019) integrate physical laws described by partial differential equations (PDEs) directly into neural network frameworks and were proposed as a framework for solving forward and inverse problems involving nonlinear PDEs~\cite{raissi2019}. Unlike traditional machine learning approaches, PINNs leverage the governing equations of physical phenomena as constraints during the training process, enhancing model interpretability, accuracy, and efficiency.

A typical PINN formulation involves embedding PDEs, initial conditions, and boundary conditions into the loss function of the neural network. The loss function comprises two components: Data Loss and Physics Loss, where the former ensures the network output aligns with available observational data utilizing a mean square error (MSE) between the networks' predictions and observational data, and the latter penalizes deviations from the governing physical equations. PINNs thus aim to minimize a composite loss function:
\begin{equation}
\mathcal{L} = \mathcal{L}_{\mathcal{D}} + \mathcal{L}_{\mathcal{P}} \label{eqn:PINNloss}
\end{equation}
where \(\mathcal{L}_{\mathcal{D}}\) and \(\mathcal{L}_{\mathcal{P}}\) denote the data loss and the physics loss respectively.

\begin{table}[tp]
\caption{Parameters of the PoB System}
\label{tab:syspar}
\centering
\footnotesize
\begin{tabular}{@{}llc@{}}
\toprule
\textbf{Parameter} & \textbf{Description} & \textbf{Value/Unit} \\ \midrule
$m_b$   & Mass of the box                  & 0.5 kg        \\
$m_L$   & Mass of the link                 & 1 kg          \\
$L$     & Length of the link               & 0.5 m         \\
$d$     & Distance from pivot to link CoM  & 0.25 m        \\
$J_L$   & Moment of inertia of the link    & 0.042 kg$\cdot$m$^2$ \\ \bottomrule
\end{tabular}
\end{table}

\subsection{Pendulum-on-a-Box System}\label{sec:PoB}
We introduce the pendulum-on-a-box (PoB) system in Fig.~\ref{fig:PoB}, which is a modification of the cart-pole system where the pole is powered instead of the cart, and the cart wheels are removed. The system must leverage surface friction by swinging the arm to achieve locomotion. We chose this system as a testbed for our framework since it shows properties similar to more challenging problems in robotics and locomotion. The system is non-linear and underactuated, and the normal force constantly fluctuates during phases of locomotion, while friction is paramount in achieving locomotion. The Lagrangian method is used to derive the equations of motion \eqref{eqn:eomx}-\eqref{eqn:eomtheta} and put into manipulator equations of the form in~\eqref{eqn:manip}.

\begin{figure*}[t]
    \centering
    \subfloat[]{%
        \includegraphics[width=0.37\textwidth]{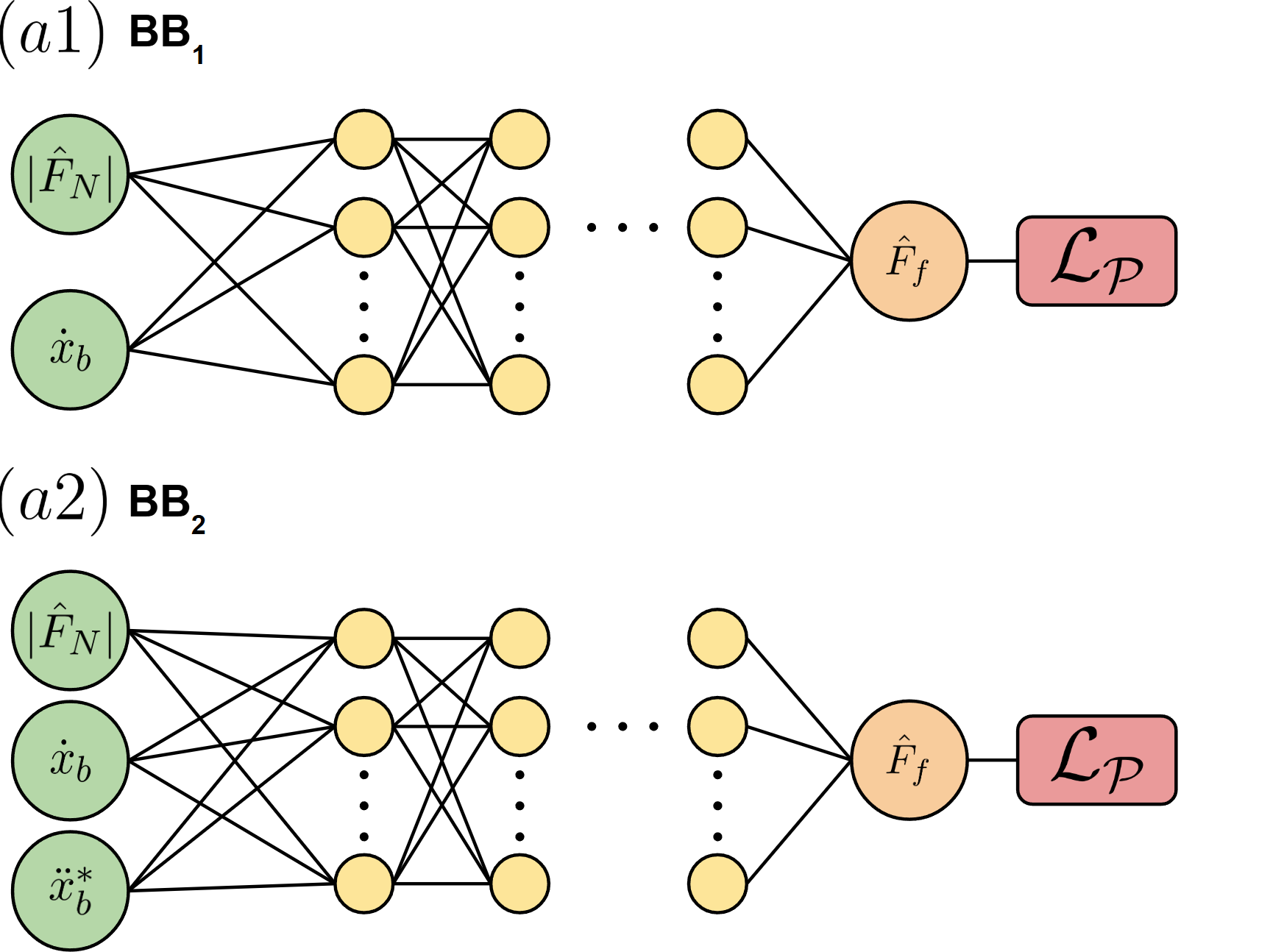}%
        \label{fig:blackboxNN}}
    \hspace{.8cm}
    \subfloat[]{%
        \includegraphics[width=0.37\textwidth]{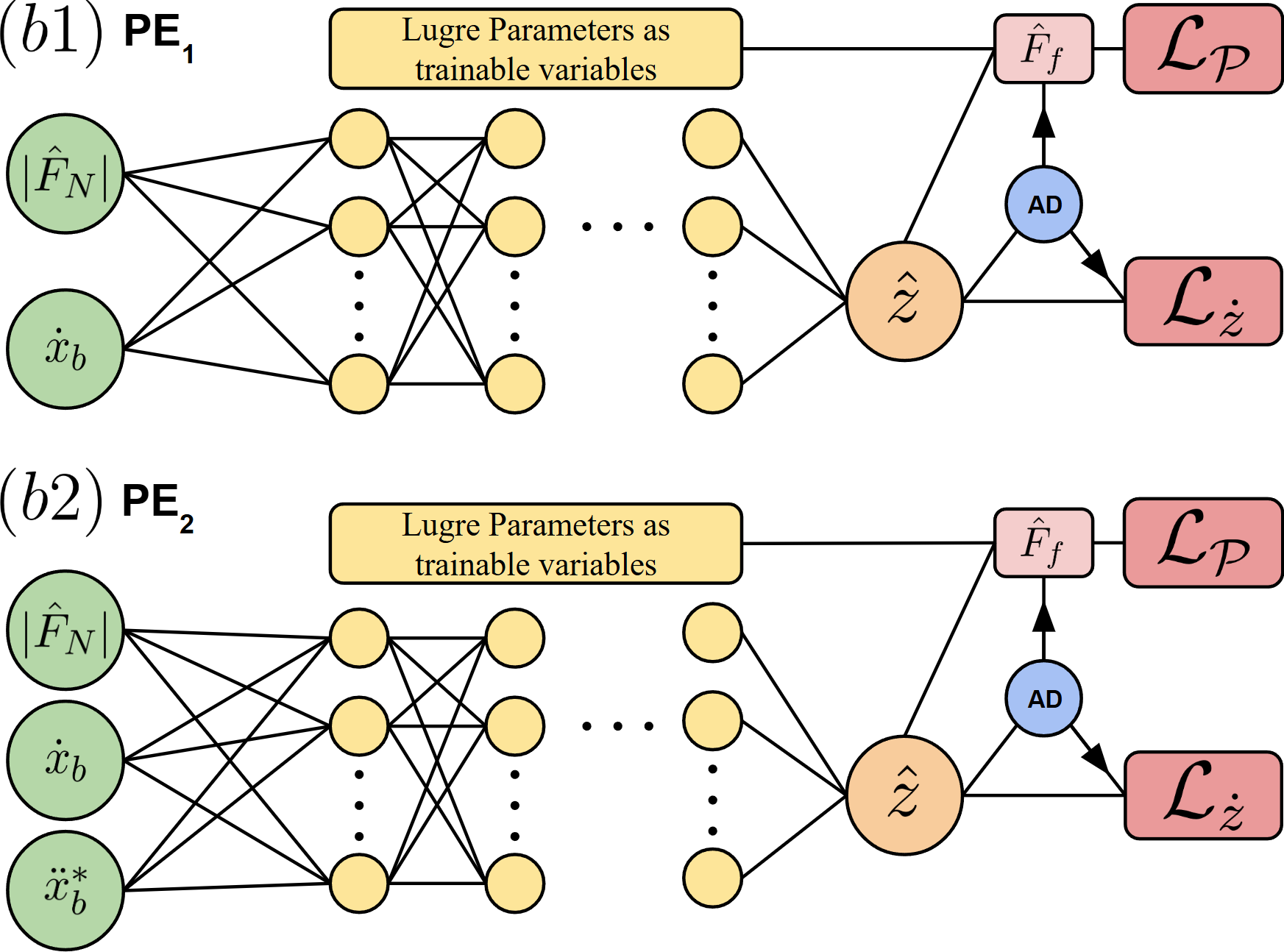}%
        \label{fig:paramEstNN}}
    \caption{I/O of Blackbox Models (a), Parameter Estimation Models (b) and corresponding loss terms. Automatic differentiation (AD) is used to compute $\hat{\dot{z}}_{model}$}
    \label{fig:NNs}
\end{figure*}
\begin{align}
    &(m_b+m_L)\ddot{x}_b = - m_Ld(\ddot{\theta}\cos\theta - \dot{\theta}^2\sin\theta) - F_f \label{eqn:eomx} \\
    &(m_b+m_L)(\ddot{y}_b + g) = m_Ld(\ddot{\theta}\sin\theta + \dot{\theta}^2\cos\theta) \hfill \label{eqn:eomy} \\
    &(J_L+m_Ld)\ddot{\theta} = - m_Ld \big((\ddot{x}_b - \dot{y}_b\dot{\theta} + \dot{y}_b)\cos\theta -\notag \\ 
    & \hspace{95pt}(\dot{x}_b\dot{\theta} + \ddot{y}_b  + \dot{x}_b  - g)\sin\theta \big) + \tau \label{eqn:eomtheta}
\end{align}
where the states $x_b$, $y_b$, $\theta$ are defined as the x and y position of the box and the angle of the link in radians, respectively. 

\begin{equation}
    \textbf{M}(\textbf{q})\ddot{\textbf{q}} + \textbf{C}(\textbf{q},\dot{\textbf{q}})\dot{\textbf{q}} + \textbf{T}_g(\textbf{q})g = \textbf{B}\textbf{u} \label{eqn:manip}
\end{equation}
where, $\textbf{q} = [x_b,y_b,\theta,\dot{x}_b,\dot{y}_b, \dot{\theta}]^T$, $\textbf{u} = [-F_f, 0,\tau]^T$, updating our definition of $\boldsymbol{\dot{z}}$ to \eqref{eqn:zdotPoB}. 
\begin{equation}
    \dot{z} = \dot{x}_b - \frac{\sigma_0 |\dot{x}_b|}{F_c + (F_s - F_c) e^{-(|\dot{x}_b|/v_s)^\alpha}} z \label{eqn:zdotPoB}
\end{equation}
The updated ${\dot{z}}$ is then used to calculate the friction force using \eqref{eqn:Ffriction} and is incorporated into the equations of motion to simulate the system. The numerical values of the parameters for the PoB system are listed in \cref{tab:syspar}. Data generation for training using this system will be described in more detail in the next section.
\section{Methods} \label{sec:methods}
We propose two different neural networks for learning generalizable friction models without \textit{a priori} knowledge of friction characteristics in the data. Henceforth, for our PINN approach, the data loss term $\mathcal{L}_\mathcal{D}$ in \eqref{eqn:PINNloss} is dropped. Our first approach is a blackbox (BB) friction model in \cref{fig:blackboxNN} that estimates the friction force directly, and the second is a parameter estimation (PE) model in \cref{fig:paramEstNN} that estimates the internal LuGre state, $z$, along with the LuGre parameters. We use feed-forward neural networks with fully connected layers for all approaches.

\subsection{PINNs for Learning Blackbox Friction Estimators}
Our approach is based on PINNs to learn generalizable friction models. The friction network BB$_1$ in Fig.~\ref{fig:NNs}(a, top), as inputs, takes the relative velocity between surfaces ${\dot{x}_b}$ and the absolute value of the normal force estimate ${\hat{F}_N}$. Given that directly measuring the normal force is not always practical, we utilize the equations of motion to estimate the normal force based on the system's states as outlined in \eqref{eqn:Fnhat} with the assumption that the system does not break contact with the surface. 
\begin{equation}
    \hat{F}_N = m_Ld\dot{\theta}^2\cos\theta - (m_b+m_L)g \label{eqn:Fnhat}
\end{equation}
The physics loss function is based on \eqref{eqn:eomx} in the equations of motion for the PoB system. 
\begin{equation}
    \mathcal{L}_\mathcal{P} = ||(m_b+m_L)\ddot{x}_b + m_Ld(\ddot{\theta}\cos\theta - \dot{\theta}^2\sin\theta) + \hat{F}_f||^2 \label{eqn:physloss}
\end{equation}

This initial setup is sufficient to be used as friction models in simulation for any system. However, we also introduce an additional neural network BB$_2$ in Fig.~\ref{fig:NNs}(a, bottom) to enable our approach to be feasible for online friction estimation. We provide an additional input to the neural network that is defined as the ``would be acceleration" of the contact point if no friction was present. In the PoB system, this equates to the would-be box acceleration in zero friction conditions, which is defined by \eqref{eqn:wouldBd2x}. 
\begin{equation}
\ddot{x}_b^* = m_Ld(-\ddot{\theta}\cos(\theta) + \dot{\theta}^2\sin\theta)/(m_b+m_L) \label{eqn:wouldBd2x}
\end{equation}
This input is derived from the system states and known EoMs and provides the network with the necessary information for online friction estimation. Inputs and outputs of these networks are outlined in \cref{fig:blackboxNN}.

\subsection{PINNs for LuGre State and Parameter Estimation}
Our second approach learns the internal state of the LuGre friction model and the underlying parameters that make it up. The physics loss in this approach includes the estimated friction force in the form of LuGre friction and an additional term $\mathcal{L}_{\dot{z}}$ with a scaling factor $\lambda$ that is used to drive the learning towards $\hat{z}$ and $\hat{\dot{z}}_{\text{model}}$ terms that are consistent with the LuGre structure.
\begin{align}
\mathcal{L} &= \mathcal{L}_\mathcal{P} + \lambda\mathcal{L}_{\dot{z}} \label{eqn:totalloss} \\
\mathcal{L}_{\dot{z}} &= ||\hat{\dot{z}}_{\text{LuGre}} - \hat{\dot{z}}_{\text{model}}||^2 \\
\hat{F}_f &= \hat{\sigma}_0 \hat{z} + \hat{\sigma}_1 \hat{\dot{z}}_{\text{model}} + \hat{\sigma}_2 \dot{x}_b \\
\hat{F}_c &= \hat{\mu}_c |\hat{F}_N| \\
\hat{F}_s &= \hat{\mu}_s |\hat{F}_N| \\
\hat{\dot{z}}_{\text{LuGre}} &= \dot{x}_b - \frac{\hat{\sigma}_0 |\dot{x}_b|}{\hat{F}_c + (\hat{F}_s - \hat{F}_c) e^{-(|\dot{x}_b|/\hat{v}_s)^2}} \hat{z}
\end{align}

In this approach, the neural network PE$_1$ in~\cref{fig:NNs}(b, top) has output $\hat{z}$ and $\hat{\dot{z}}_{\text{model}}$ denotes its derivative. $\hat{\sigma}_{0,1,2}$, $\hat{\mu}_{c,s}$ and $\hat{v}_s$ are estimated parameters of the LuGre friction model; ${\alpha}$ in \eqref{eqn:zdot} is set to 2 as it is commonly adopted. $\hat{\dot{z}}_{\text{LuGre}}$ is an estimate of $\dot{z}$ from LuGre formulation in \eqref{eqn:zdotPoB} using the $\hat{z}$, and estimated LuGre parameters. The LuGre parameters to be estimated are added to the neural network as trainable variables to be learned and adjusted during training. Similar to the blackbox networks, $\ddot{x}_b^*$ is used as an additional input for PE$_2$ in the secondary approach in~\cref{fig:NNs}(b, bottom) for parameter estimation models to enable online friction estimation where necessary. %This approach also enables the possibility of using the learned parameters as a traditional LuGre friction model if a hybrid learned friction model is not preferred\\
\begin{figure}[tbp]
    \centering
    \includegraphics[width=\columnwidth]{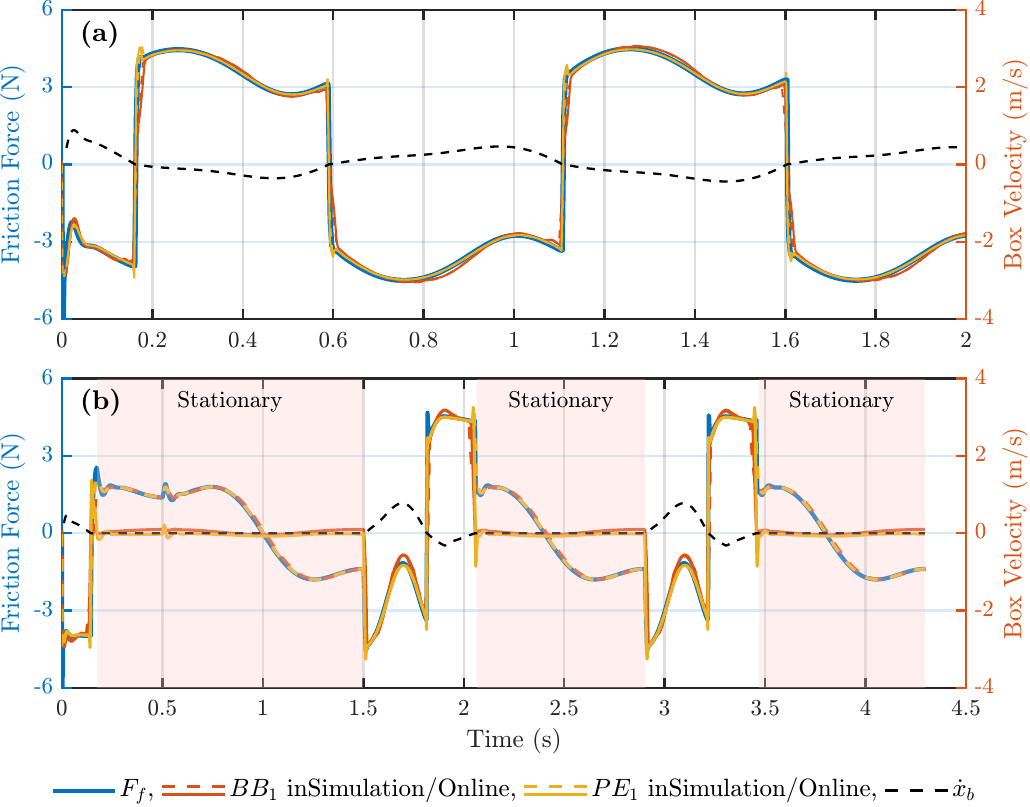}
    % \caption{\footnotesize BB$_1$ and PE$_1$ friction estimations for: \textbf{(a)}  Traj. 1, \textbf{(b)}  Traj. 2}  
    \caption{\footnotesize BB$_1$ and PE$_1$ friction estimations for: \textbf{(a)} Traj.~1, \textbf{(b)} Traj.~2. Overall MSE (N$^2$) for in-simulation / online: \textbf{(a)} BB$_1$: 0.041 / 0.124, PE$_1$: 0.014 / 0.032; \textbf{(b)} BB$_1$: 0.092 / 1.674, PE$_1$: 0.012 / 1.591.}
    \label{fig:simpModels}
\end{figure}
\begin{figure}[bp]
    \centering
    \includegraphics[width=\columnwidth]{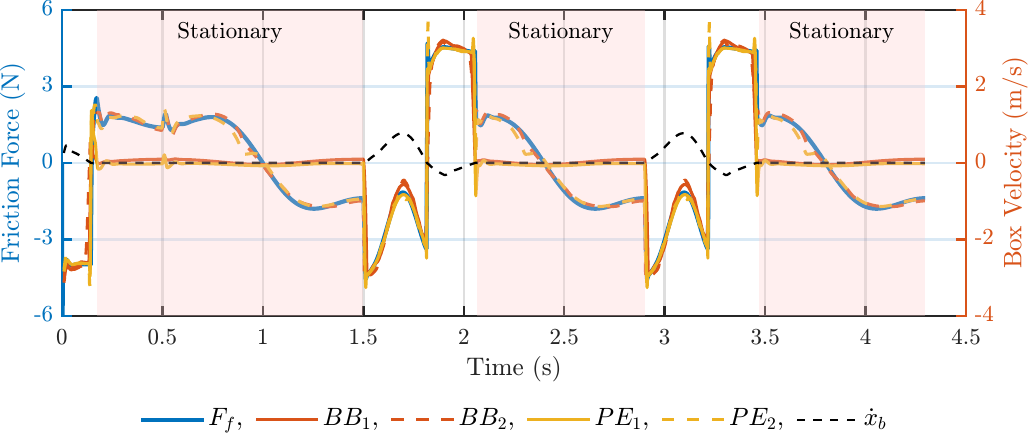}
    % \caption{\footnotesize Online estimation performances of BB$_{1,2}$ and PE$_{1,2}$ for Traj. 2.} 
    \caption{\footnotesize Online estimation performances of BB$_{1,2}$ and PE$_{1,2}$ for Traj.~2. Overall MSE (N$^2$): BB$_1$: 1.674, BB$_2$: 0.174, PE$_1$: 1.591, PE$_2$: 0.066.}
    \label{fig:acc_models}
\end{figure}
\subsection{Data Generation and Training}
% Training data for our PINN method is created using MATLAB R2023a, where the system dynamics are simulated by numerically solving the equations of motion using ode45. The LuGre model defined by \eqref{eqn:Ffriction}, \eqref{eqn:zdotPoB} was selected as the ground truth for testing due to its complexity and widespread use in a variety of fields.
% Training data for the networks consists of 6 different trials sampled at 400Hz, equating to approximately 5800 points and 14.5 seconds of data. 5 of these datasets are of the pendulum swinging at the same frequency with swing ranges from $\pm$35 to $\pm$65 degrees, each consisting of two seconds of data. The last dataset is the PoB system moving in the +x direction by swinging the pendulum using a custom trajectory. A Gaussian noise of 5\% of the signals' standard deviation is added to each corresponding signal to simulate sensor noise. These short trajectories with noise are selected for the training data to show that our approach works with minimal generic data. The networks are then trained using the noisy dataset via TensorFlow 2.15.0 in Python 3.10.12.
Training data for the PINN is generated in MATLAB R2023a by simulating the system dynamics with \texttt{ode45}. The LuGre model in \eqref{eqn:Ffriction} and \eqref{eqn:zdotPoB} is used as ground truth due to its complexity and broad adoption.

Six trials were collected at 400 Hz, yielding about 5800 samples over 14.5 s. Five trials are pendulum swings at the same excitation frequency with amplitudes ranging from $\pm35^{\circ}$ to $\pm65^{\circ}$, each lasting 2 s. The sixth trial is a PoB translation in $+x$ using a custom swing trajectory. Gaussian noise at 5\% of each signal’s standard deviation is added to simulate sensor noise. These short, noisy trajectories are chosen to demonstrate learning from minimal generic data. Networks are trained on the noisy set using TensorFlow 2.15.0 in Python 3.10.12. Architectures are as follows: BB$_{1}$ and PE$_{1}$ use four hidden layers with 128 neurons per layer, and BB$_{2}$ and PE$_{2}$ use four hidden layers with 512 neurons per layer.

All networks use the Adam optimizer with ReLU activations, and are trained on the full batch for 10,000 epochs. BB$_{1}$ and PE$_{1}$ use a fixed learning rate of $1\!\times\!10^{-4}$. BB$_{2}$ and PE$_{2}$ use an initial learning rate of $1\!\times\!10^{-3}$ with adaptive learning rate scheduling that halves the rate when the loss plateaus, down to a minimum of $1\!\times\!10^{-5}$. For the parameter estimation models PE$_{1,2}$, the scaling factor in \eqref{eqn:totalloss} is set to $\lambda = 1\!\times\!10^{5}$ to balance the EoM residual $\mathcal{L}_\mathcal{P}$ against the bristle deflection rate consistency loss $\mathcal{L}_{\dot{z}}$, which operate at different magnitudes.
\begin{figure}[tp]
    \centering
    \includegraphics[width=\columnwidth]{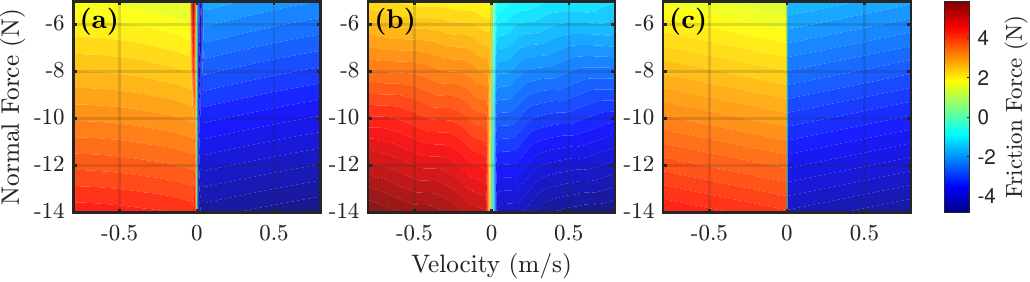}
    \caption{\footnotesize Friction force characteristics for \textbf{(a)} PE$_1$, \textbf{(b)} BB$_1$, \textbf{(c)} LuGre model}
    \label{fig:steady_state}
\end{figure}
\begin{figure}[bp]
    \centering
    \includegraphics[width=0.5\columnwidth]{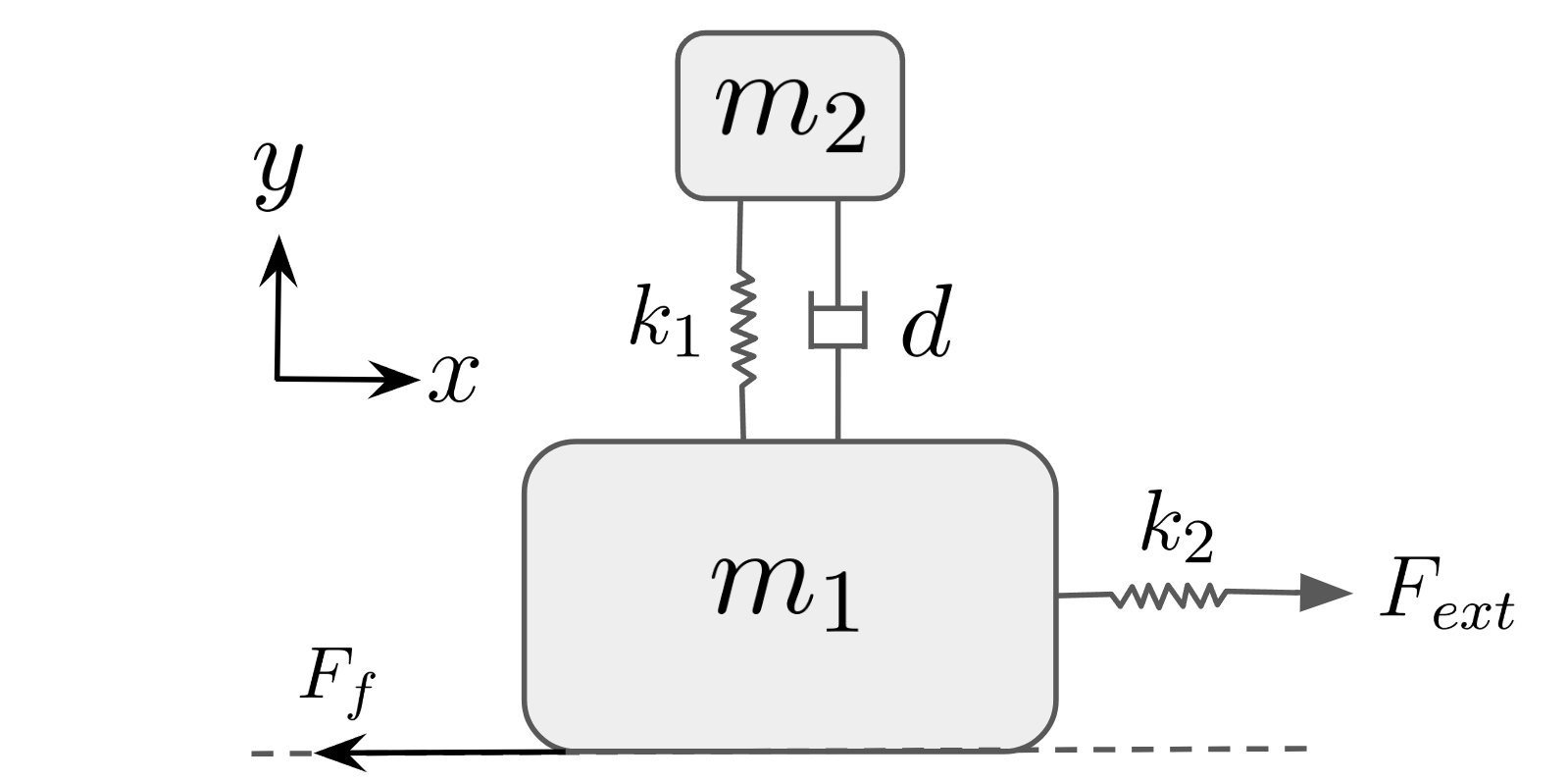}
    \caption{\footnotesize Spring-Damper on a Box (\textit{SDoB}) system}
    \label{fig:SDoB_FoB}
\end{figure}
\section{Results and Discussion}\label{sec:results}
% In this section, we show that simple PINN-inspired friction estimators trained on only velocity and normal force data (BB$_1$ and PE$_1$) are sufficient to create friction estimators that can be used in simulation. We also discuss the limitations of this approach and why an additional approach is needed to use PINNs as online friction estimators (BB$_2$, PE$_2$), and present that our approach results in transferable learned friction models. Lastly, we argue that this approach enables the use of more accurate friction models than simple viscous or coulombic+viscous friction models with minimal data collection and training, and it is quicker to deploy than running a full LuGre parameter estimation.
We evaluate the proposed friction estimators for in-simulation use, online estimation, transferability across different dynamical systems, and LuGre parameter identification speed and accuracy.
\begin{table*}[t]
\caption{Estimated LuGre Friction Parameters}
\label{tab:parEst}
\centering
\footnotesize
\newcommand{\pu}[2]{#1\hfill{\scriptsize(#2)}} % parameter + right-aligned unit in col 1
\begin{tabular}{@{}lccccccc@{}}
\toprule
\textbf{Pars.} & \textbf{Ground Truth} & \textbf{PE$_1$} & \textbf{PE$_2$} & \textbf{Nelder–Mead} & \textbf{Genetic Alg.} & \textbf{Nonlinear LS} \\
\midrule
\pu{$\sigma_0$}{N/m}     & $1.00\!\times\!10^5$    & $1.20\!\times\!10^5$       & $1.02\!\times\!10^5$    & $1.03\!\times\!10^5$   & $1.04\!\times\!10^5$   & $99.9\!\times\!10^5$  \\
\pu{$\sigma_1$}{Ns/m}    & 316.23                   & 346.41                   & 319.37                   & 320.36  & 322.43  & 316.21   \\
\pu{$\sigma_2$}{Ns/m}    & 0.40                     & 0.41                     & 0.44                     & 0.20 & 0.47 & 1        \\
$\mu_c$           & 0.30                     & 0.28                     & 0.30                     & 0.30 & 0.30 & 0.23 \\
$\mu_s$           & 0.60                     & 0.60                     & 0.61                     & 0.49 & 0.59 & 0.47 \\
\pu{$v_s$}{m/s}          & $10.0\!\times\!10^{-4}$  & $8.00\!\times\!10^{-4}$   & $9.96\!\times\!10^{-4}$  & $4.13\!\times\!10^{-4}$ & $11.9\!\times\!10^{-4}$ & $57.9\!\times\!10^{-4}$ \\
\pu{$t_{comp}$}{min}     & -                        & 9.46                      & 10.54                     & 15.18                    & 67.80                    & 24.14 \\
\bottomrule
\end{tabular}
\end{table*}
\subsection{PINN Friction Models for Use in Simulation}
We evaluate BB$_1$ and PE$_1$ models as in-simulation (dashed) friction models and online friction estimators (solid) in \cref{fig:simpModels}. In the first approach, the friction force estimates from the trained neural networks are incorporated into \eqref{eqn:eomx}, and the PoB system is simulated using the EoMs outlined in \eqref{eqn:eomx}-\eqref{eqn:eomtheta} and the differential equations are solved using ode45 in MATLAB. In the second approach, the data collection is done with the ground truth LuGre friction model, and the noisy data at each timestep is fed into the trained neural networks to estimate the friction force at that time step for online estimation of the friction force. We test the models on two different trajectories: a constant oscillation of the pendulum at a set range and a custom trajectory designed to move the PoB system in the +x direction by swinging the pendulum at different speeds to leverage the stick and slip nature of friction between the surface and the PoB system. For the sake of simplicity, in the rest of this paper we will refer to these trajectories as ``Traj.~1'' and ``Traj.~2'' respectively.

The trained friction models BB$_1$ and PE$_1$ appear to perform well in both in-simulation and online estimation for Traj.~1 in \cref{fig:simpModels}a. However, in trajectories similar to Traj.~2 (\cref{fig:simpModels}b), where there are extended stationary periods during which the friction force depends on the internal forces of the system, their online estimation capabilities fall short. In these trajectories, the inputs to the models BB$_1$ and PE$_1$ ($\dot{x}_b$,$F_N$) do not provide sufficient information for estimating friction and therefore fail in the stationary regime when used as online estimators. PE$_1$ achieves the lowest in-simulation MSE across both trajectories (0.014 and 0.012~N$^2$), while online estimation errors on Traj.~2 are significantly higher (BB$_1$: 1.674, PE$_1$: 1.591~N$^2$) due to insufficient information during stationary periods.

% Dynamic friction models that depend on velocity and normal force, like the LuGre model, rely on micro-motions that happen after a numerical integration step to deflect the `bristles' and generate a reactive force to the object on the surface. This reactive friction force builds due to the bristle deflection, even though the object would seem stationary to an observer. Due to these micro-motions, the LuGre friction model can capture friction behavior just from velocity and normal force ($\dot{x}_b$,$F_N$).

% Similarly, our trained friction models, BB$_1$ and PE$_1$, can accurately capture friction phenomena when used as in-simulation models. These micro-motions result in an accurate friction estimation and render this approach feasible for learning friction models for in-simulation use but not as online estimators. We see this as a limitation of this approach as online friction estimation is important for deploying systems in the real world and propose a secondary approach for online estimation using PINNs.
Dynamic friction models like LuGre rely on micro-motions after each numerical integration step to deflect the bristles and generate a reactive force, even when the object appears stationary. Our trained models BB$_1$ and PE$_1$ similarly capture these phenomena when used in simulation, but this mechanism is unavailable during online estimation. We see this as a limitation and propose a secondary approach for online estimation using PINNs.

\subsection{PINN Friction Models for Online Friction Estimation}
Friction estimation is an important part of planning trajectories and tracking accuracy; hence, our proposed secondary models, BB$_2$ and PE$_2$, are improvements upon BB$_1$ and PE$_1$ in that they can be used to estimate friction on the go. The additional input of the secondary models enables them to break free of the limitations that BB$_1$ and PE$_1$ suffer from by having only velocity and normal force information and removes the need to simulate the dynamics for reliable friction estimation. \cref{fig:acc_models} compares the online estimation performance of all the PINN-based friction estimation models. The 3-input models substantially reduce online estimation error on Traj.~2: BB$_2$ and PE$_2$ achieve MSEs of 0.174 and 0.066~N$^2$ (reductions of 90\% and 96\% over their 2-input counterparts).
\begin{figure}[bp]
    \centering
    \includegraphics[width=\columnwidth]{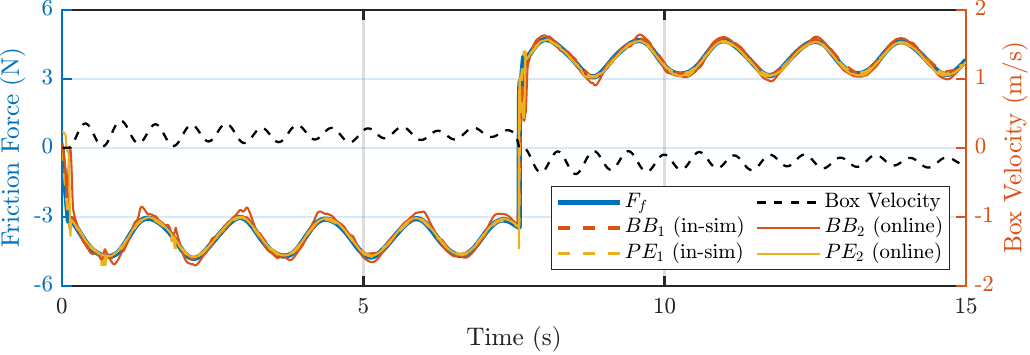}
    \caption{\footnotesize Models trained on PoB for friction estimation in the SDoB system.}
    \label{fig:SDoB_comparison}
\end{figure}
\subsection{Transferability of Learned Friction Models and Limitations}
Our approach enables the learned models to be transferable to different systems that are deployed in the same environment. Learned models exhibit behavior similar to the LuGre model for different velocity and normal force pairs (\cref{fig:steady_state}), enabling them to be used on different systems than the ones they are trained on if the system is deployed in the same environment. To test this, we introduce a system called Spring-Damper on a Box (\textit{SDoB}) illustrated in Fig.~\ref{fig:SDoB_FoB}, consisting of two masses connected by a spring and a damper pulled by an external force applied on the bottom mass ($m_1$). The top mass ($m_2$) in this system is free to move in the y direction, which results in varying normal force throughout the overall system trajectory. We simulated the \textit{SDoB} system using the same underlying LuGre friction model as the \textit{PoB} system and tested the learned friction models that are trained on the \textit{PoB} system to estimate friction on the SDoB system (Fig.~\ref{fig:SDoB_comparison}). Our results strongly suggest that the training framework proposed in this paper enables the trained models to be transferable to different systems or to the same dynamic system with different parameters. However, the extent of this transferability is inherently limited: significant modifications to the underlying system properties, such as changes in stiffness or flexibility, or the introduction of new materials at the contact interface with different frictional characteristics, may reduce model accuracy and require re-training or adaptation.
\begin{figure}[bp]
    \centering
    \includegraphics[width=\columnwidth]{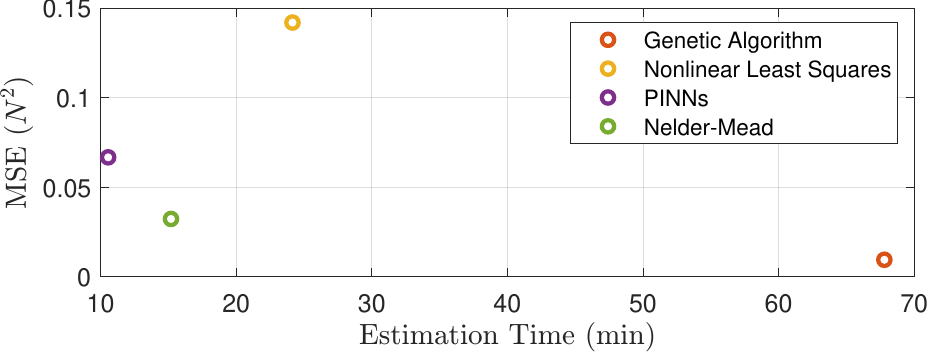}
    \caption{\footnotesize Comparison of various parameter estimation methods with respect to computation time and mean squared error (MSE) of friction force using the estimated parameters.}
    \label{fig:speedvsacc}
\end{figure}
\subsection{Parameter Estimation Performance}
Choosing an appropriate friction model for each setup is an important decision. The complexity of the friction model, the identification process for parameters, the time it takes, and the computational cost are all essential decision factors in deciding which friction model is the best for a particular setup. Although the LuGre friction model accurately captures most aspects of the actual frictional behavior, such as stick-slip, it requires more work to identify the parameters that fit the experimental data. %The dynamic nature of the LuGre friction model requires the internal state to be estimated via the differential equations that define the system and the LuGre friction model. Due to the high resolution required to capture the transition regions (stick-slip), variable step solvers like ode45 are required. This lends itself to high computational cost and high computation times, often resulting in simpler friction models being chosen.%

We show that our methods employing PINNs enable faster identification of friction models of comparable complexity with sufficient accuracy. \cref{tab:parEst} compares the parameter estimation performance of established methods with our two models $PE_{1,2}$ shown in \cref{fig:paramEstNN}. The performance of our models is on par with the other methods. A performance trade-off between mean squared error (MSE) of the estimated friction forces and computation time is illustrated in \cref{fig:speedvsacc}. While our approach may not replace full parameter identification in safety-critical systems where maximum accuracy is essential, it provides a faster alternative for estimating LuGre parameters. Although Nelder-Mead simplex method appears competitive in \cref{fig:speedvsacc}, its parameter estimation accuracy was worse (\cref{tab:parEst}). 
\section{Conclusion and Future Work}
% We formulated a PINN-based friction estimation framework for creating transferable learned friction models. We showed the efficacy of these models both as in-simulation models and online friction estimators. We also presented the transferability of these models on a system that they were not trained on. We then discussed our framework's computational speed and accuracy in parameter identification of the LuGre friction model compared to other methods and the appeal of this approach as a fast way of shrinking the sim-to-real gap. Future work includes incorporating this framework with existing control methods, trajectory optimization, and expanding the framework to accommodate online adaption to changing friction properties. In each such scenario, we anticipate a Pareto trade-off of the same general form (\cref{fig:speedvsacc}), with the exact trade-offs between computational speed and accuracy of motion being a key concern for future work.
We formulated a PINN-based friction estimation framework for creating transferable learned friction models. We demonstrated the efficacy of these models both as in-simulation friction models and online friction estimators, showed their transferability to a dynamical system they were not trained on, and compared the framework's computational speed and accuracy in LuGre parameter identification against established methods.

The present results are validated using synthetic data generated from the LuGre model. While this demonstrates that the framework can recover complex friction behavior from minimal state measurements, experimental validation on physical hardware is needed to fully assess its applicability to real-world systems. Further work includes extending the framework to multi-contact and higher degree-of-freedom systems, systematic comparison with other hybrid learning-based friction modeling approaches, and integration with existing control methods and trajectory optimization. Expanding the framework to accommodate online adaptation to changing friction properties is also planned. In each such scenario, we anticipate a Pareto trade-off of the same general form (\cref{fig:speedvsacc}), with the exact trade-offs between computational speed and accuracy of motion being a key concern.
% \addtolength{\textheight}{-14.3cm}
\bibliographystyle{IEEEtran}
\bibliography{references}

@article{canudas1995new,
  title={A new model for control of systems with friction},
  author={Canudas de Wit, Carlos and Olsson, H{\aa}kan and {\AA}str{\"o}m, Karl Johan and Lischinsky, Mattias},
  journal={IEEE Transactions on Automatic Control},
  volume={40},
  number={3},
  pages={419--425},
  year={1995},
  publisher={IEEE}
}

@article{astrom2008revisiting,
  title={Revisiting the {LuGre} friction model},
  author={{\AA}str{\"o}m, Karl J and Canudas-de-Wit, Carlos},
  journal={IEEE Control Systems Magazine},
  volume={28},
  number={6},
  pages={101--114},
  year={2008},
  publisher={IEEE}
}

@article{rillLuGreNotLuGre2024,
  title = {{{LuGre}} or Not {{LuGre}}},
  author = {Rill, G. and Schaeffer, T. and Schuderer, M.},
  date = {2024-02-01},
  journaltitle = {Multibody System Dynamics},
  shortjournal = {Multibody Syst Dyn},
  volume = {60},
  number = {2},
  pages = {191--218},
  issn = {1573-272X},
  doi = {10.1007/s11044-023-09909-5},
  url = {https://doi.org/10.1007/s11044-023-09909-5},
  urldate = {2024-02-18},
  langid = {english},
}

@proceedings{cyrusID,
    author = {Sun, Yun-Hsiang and Chen, Tao and Shafai, Cyrus},
    title = {Parameter Identification of LuGre Friction Model: Experimental Set-up Design and Measurement},
    volume = {Volume 4A: Dynamics, Vibration, and Control},
    series = {ASME International Mechanical Engineering Congress and Exposition},
    pages = {V04AT04A057},
    year = {2015},
    month = {11},
    doi = {10.1115/IMECE2015-51255},
    url = {https://doi.org/10.1115/IMECE2015-51255},
    eprint = {https://asmedigitalcollection.asme.org/IMECE/proceedings-pdf/IMECE2015/57397/V04AT04A057/4267302/v04at04a057-imece2015-51255.pdf},
}

@article{raissi2019,
  title={Physics-informed neural networks: A deep learning framework for solving forward and inverse problems involving nonlinear partial differential equations},
  author={Raissi, Maziar and Perdikaris, Paris and Karniadakis, George Em},
  journal={Journal of Computational Physics},
  volume={378},
  pages={686--707},
  year={2019},
  publisher={Elsevier}
}

@article{karniadakis2021,
  title={Physics-informed machine learning},
  author={Karniadakis, George Em and Kevrekidis, Ioannis G and Lu, Lu and Perdikaris, Paris},
  journal={Nature Reviews Physics},
  volume={3},
  number={6},
  pages={422--440},
  year={2021},
  publisher={Springer}
}

@article{li2024,
  title={Physics-informed neural networks for friction-involved nonsmooth dynamics problems},
  author={Li, Zilin and Bai, Jinshuai and Ouyang, Huajiang and others},
  journal={Nonlinear Dynamics},
  volume={110},
  pages={2345--2361},
  year={2024},
  publisher={Springer}
}

@article{coble2024,
  title={Physics-informed machine learning for dry friction and backlash modeling in structural control systems},
  author={Coble, Daniel and Cao, Liang and Downey, Austin RJ},
  journal={Mechanical Systems and Signal Processing},
  volume={218},
  pages={111522},
  year={2024},
  publisher={Elsevier}
}

@article{sorrentino2024,
  title={Physics-informed learning for the friction modeling of high-ratio harmonic drives},
  author={Sorrentino, Ines and Romualdi, Giulio and Bergonti, Fabio and others},
  journal={IEEE Robotics and Automation Letters},
  volume={9},
  pages={3500--3510},
  year={2024},
  publisher={IEEE}
}

@article{scholl2024,
  title={Learning-based adaption of robotic friction models},
  author={Scholl, Philipp and Iskandar, Maged and Wolf, Sebastian and others},
  journal={Robotics and Computer-Integrated Manufacturing},
  volume={89},
  pages={102780},
  year={2024},
  publisher={Elsevier}
}

@article{quantumChem,
  title = {Ab initio solution of the many-electron Schr\"odinger equation with deep neural networks},
  author = {Pfau, David and Spencer, James S. and Matthews, Alexander G. D. G. and Foulkes, W. M. C.},
  journal = {Phys. Rev. Res.},
  volume = {2},
  issue = {3},
  pages = {033429},
  numpages = {20},
  year = {2020},
  month = sep,
  publisher = {American Physical Society},
  doi = {10.1103/PhysRevResearch.2.033429},
  url = {https://link.aps.org/doi/10.1103/PhysRevResearch.2.033429}
}

@article{4DflowMRI,
title = {Machine learning in cardiovascular flows modeling: Predicting arterial blood pressure from non-invasive 4D flow MRI data using physics-informed neural networks},
journal = {Computer Methods in Applied Mechanics and Engineering},
volume = {358},
pages = {112623},
year = {2020},
issn = {0045-7825},
doi = {https://doi.org/10.1016/j.cma.2019.112623},
url = {https://www.sciencedirect.com/science/article/pii/S0045782519305055},
author = {Georgios Kissas and Yibo Yang and Eileen Hwuang and Walter R. Witschey and John A. Detre and Paris Perdikaris}
}

@article{flowEspresso,
   title={Flow over an espresso cup: inferring 3-D velocity and pressure fields from tomographic background oriented Schlieren via physics-informed neural networks},
   volume={915},
   ISSN={1469-7645},
   url={http://dx.doi.org/10.1017/jfm.2021.135},
   DOI={10.1017/jfm.2021.135},
   journal={Journal of Fluid Mechanics},
   publisher={Cambridge University Press (CUP)},
   author={Cai, Shengze and Wang, Zhicheng and Fuest, Frederik and Jeon, Young Jin and Gray, Callum and Karniadakis, George Em},
   year={2021},
   month=mar }

@article{plasmaPhys,
   title={Uncovering turbulent plasma dynamics via deep learning from partial observations},
   volume={104},
   ISSN={2470-0053},
   url={http://dx.doi.org/10.1103/PhysRevE.104.025205},
   DOI={10.1103/physreve.104.025205},
   number={2},
   journal={Physical Review E},
   publisher={American Physical Society (APS)},
   author={Mathews, A. and Francisquez, M. and Hughes, J. W. and Hatch, D. R. and Zhu, B. and Rogers, B. N.},
   year={2021},
   month=aug }

@misc{structural,
      title={Physics-informed neural network for ultrasound nondestructive quantification of surface breaking cracks}, 
      author={Khemraj Shukla and Patricio Clark Di Leoni and James Blackshire and Daniel Sparkman and George Em Karniadakis},
      year={2020},
      eprint={2005.03596},
      archivePrefix={arXiv},
      primaryClass={cs.LG},
      url={https://arxiv.org/abs/2005.03596}, 
}

@article{olejnik2023,
  title = {Friction Modelling and the Use of a Physics-Informed Neural Network for Estimating Frictional Torque Characteristics},
  author = {Olejnik, Paweł and Ayankoso, Samuel},
  date = {2023-10-01},
  journaltitle = {Meccanica},
  shortjournal = {Meccanica},
  volume = {58},
  number = {10},
  pages = {1885--1908},
  issn = {1572-9648},
  doi = {10.1007/s11012-023-01716-8},
  url = {https://doi.org/10.1007/s11012-023-01716-8},
  urldate = {2024-01-31},
  langid = {english},
}

@misc{le2024contact,
      title={Contact Models in Robotics: a Comparative Analysis}, 
      author={Quentin Le Lidec and Wilson Jallet and Louis Montaut and Ivan Laptev and Cordelia Schmid and Justin Carpentier},
      year={2024},
      eprint={2304.06372},
      archivePrefix={arXiv},
      primaryClass={cs.RO},
      url={https://arxiv.org/abs/2304.06372}, 
}

@INPROCEEDINGS{iGibson,
  author={Shen, Bokui and Xia, Fei and Li, Chengshu and Martín-Martín, Roberto and Fan, Linxi and Wang, Guanzhi and Pérez-D’Arpino, Claudia and Buch, Shyamal and Srivastava, Sanjana and Tchapmi, Lyne and Tchapmi, Micael and Vainio, Kent and Wong, Josiah and Fei-Fei, Li and Savarese, Silvio},
  booktitle={2021 IEEE/RSJ International Conference on Intelligent Robots and Systems (IROS)}, 
  title={iGibson 1.0: A Simulation Environment for Interactive Tasks in Large Realistic Scenes}, 
  year={2021},
  volume={},
  number={},
  pages={7520-7527},
  doi={10.1109/IROS51168.2021.9636667}}

@InProceedings{airSim,
author="Shah, Shital
and Dey, Debadeepta
and Lovett, Chris
and Kapoor, Ashish",
editor="Hutter, Marco
and Siegwart, Roland",
title="AirSim: High-Fidelity Visual and Physical Simulation for Autonomous Vehicles",
booktitle="Field and Service Robotics",
year="2018",
publisher="Springer International Publishing",
address="Cham",
pages="621--635",
isbn="978-3-319-67361-5"
}

@ARTICLE{contactModelsinSim,
  author={Horak, Peter C. and Trinkle, Jeff C.},
  journal={IEEE Robotics and Automation Letters}, 
  title={On the Similarities and Differences Among Contact Models in Robot Simulation}, 
  year={2019},
  volume={4},
  number={2},
  pages={493-499},
  doi={10.1109/LRA.2019.2891085}}

@Inbook{Diehl2006,
author="Diehl, M.
and Bock, H.G.
and Diedam, H.
and Wieber, P.-B.",
editor="Diehl, Moritz
and Mombaur, Katja",
title="Fast Direct Multiple Shooting Algorithms for Optimal Robot Control",
bookTitle="Fast Motions in Biomechanics and Robotics: Optimization and Feedback Control",
year="2006",
publisher="Springer Berlin Heidelberg",
address="Berlin, Heidelberg",
pages="65--93",
isbn="978-3-540-36119-0",
doi="10.1007/978-3-540-36119-0_4",
url="https://doi.org/10.1007/978-3-540-36119-0_4"
}

@article{mpc2014,
title = {Model predictive control: Recent developments and future promise},
journal = {Automatica},
volume = {50},
number = {12},
pages = {2967-2986},
year = {2014},
issn = {0005-1098},
doi = {https://doi.org/10.1016/j.automatica.2014.10.128},
url = {https://www.sciencedirect.com/science/article/pii/S0005109814005160},
author = {David Q. Mayne}
}

@INPROCEEDINGS{nonlinearMPC,
  author={Kleff, Sebastien and Meduri, Avadesh and Budhiraja, Rohan and Mansard, Nicolas and Righetti, Ludovic},
  booktitle={2021 IEEE International Conference on Robotics and Automation (ICRA)}, 
  title={High-Frequency Nonlinear Model Predictive Control of a Manipulator}, 
  year={2021},
  volume={},
  number={},
  pages={7330-7336},
  doi={10.1109/ICRA48506.2021.9560990}}

@ARTICLE{mpc2022,
  author={Dantec, Ewen and Taïx, Michel and Mansard, Nicolas},
  journal={IEEE Robotics and Automation Letters}, 
  title={First Order Approximation of Model Predictive Control Solutions for High Frequency Feedback}, 
  year={2022},
  volume={7},
  number={2},
  pages={4448-4455},
  doi={10.1109/LRA.2022.3149573}}

@article{marques2021,
  title = {An Investigation of a Novel {{LuGre-based}} Friction Force Model},
  author = {Marques, Filipe and Woli{\'n}ski, {\L}ukasz and Wojtyra, Marek and Flores, Paulo and Lankarani, Hamid M.},
  year = {2021},
  month = dec,
  journal = {Mechanism and Machine Theory},
  volume = {166},
  pages = {104493},
  issn = {0094114X},
  doi = {10.1016/j.mechmachtheory.2021.104493},
  urldate = {2024-01-05},
  langid = {english},
}
\end{document}